\documentclass[11pt]{article}

\usepackage[margin=1in]{geometry}
\usepackage[T1]{fontenc}
\usepackage[utf8]{inputenc}
\usepackage{amsmath}
\usepackage{amssymb}
\usepackage{booktabs}
\usepackage{array}
\usepackage{microtype}
\usepackage{tikz}
\usetikzlibrary{arrows.meta}
\usepackage[colorlinks=true, linkcolor=blue, urlcolor=blue, citecolor=blue]{hyperref}

\definecolor{flowteal}{RGB}{14,138,131}
\definecolor{blockedred}{RGB}{192,59,45}

\newcommand{\bibentry}[1]{\par\noindent\hangindent=1.5em\hangafter=1 #1\par\vspace{2pt}}

\newcommand{\FSopen}[3]{\begin{scope}[shift={(#1,#2)}, opacity=#3]
  \fill[black!75, rounded corners=1pt] (0,-0.13) rectangle (0.45,0.13);
  \draw[line width=0.5pt] (0.45,0.08) -- (1.62,0.08);
  \draw[line width=0.5pt] (0.45,-0.08) -- (1.62,-0.08);
  \draw[flowteal, line width=0.9pt] (0.5,0) -- (0.86,0);
  \draw[line width=0.6pt] (1.02,0) circle (0.145);
  \draw[line width=0.6pt] (0.93,0.09) -- (1.11,-0.09);
  \draw[flowteal, line width=0.9pt, -{Stealth[length=3pt]}] (1.18,0) -- (1.58,0);
  \draw[line width=0.6pt] (1.66,0.2) -- (1.66,-0.2) -- (2.06,-0.2) -- (2.06,0.2);
  \fill[flowteal, opacity=0.5] (1.69,-0.17) rectangle (2.03,0.12);
\end{scope}}
\newcommand{\FScross}[3]{\begin{scope}[shift={(#1,#2)}, opacity=#3]
  \fill[black!75, rounded corners=1pt] (0,-0.13) rectangle (0.45,0.13);
  \draw[line width=0.5pt] (0.45,0.08) -- (1.62,0.08);
  \draw[line width=0.5pt] (0.45,-0.08) -- (1.62,-0.08);
  \draw[flowteal, line width=0.9pt] (0.5,0) -- (0.86,0);
  \draw[blockedred, line width=0.7pt] (1.02,0) circle (0.145);
  \draw[blockedred, line width=0.7pt] (0.92,-0.10) -- (1.12,0.10);
  \draw[blockedred, line width=0.7pt] (0.92,0.10) -- (1.12,-0.10);
  \draw[line width=0.6pt] (1.66,0.2) -- (1.66,-0.2) -- (2.06,-0.2) -- (2.06,0.2);
\end{scope}}
\newcommand{\FSopenM}[3]{\begin{scope}[shift={(#1,#2)}, opacity=#3]
  \draw[line width=0.6pt] (0,0.2) -- (0,-0.2) -- (0.4,-0.2) -- (0.4,0.2);
  \fill[flowteal, opacity=0.5] (0.03,-0.17) rectangle (0.37,0.12);
  \draw[line width=0.5pt] (0.44,0.08) -- (1.61,0.08);
  \draw[line width=0.5pt] (0.44,-0.08) -- (1.61,-0.08);
  \draw[flowteal, line width=0.9pt] (1.56,0) -- (1.2,0);
  \draw[line width=0.6pt] (1.04,0) circle (0.145);
  \draw[line width=0.6pt] (0.95,0.09) -- (1.13,-0.09);
  \draw[flowteal, line width=0.9pt, -{Stealth[length=3pt]}] (0.88,0) -- (0.48,0);
  \fill[black!75, rounded corners=1pt] (1.61,-0.13) rectangle (2.06,0.13);
\end{scope}}
\newcommand{\FScrossM}[3]{\begin{scope}[shift={(#1,#2)}, opacity=#3]
  \draw[line width=0.6pt] (0,0.2) -- (0,-0.2) -- (0.4,-0.2) -- (0.4,0.2);
  \draw[line width=0.5pt] (0.44,0.08) -- (1.61,0.08);
  \draw[line width=0.5pt] (0.44,-0.08) -- (1.61,-0.08);
  \draw[flowteal, line width=0.9pt] (1.56,0) -- (1.2,0);
  \draw[blockedred, line width=0.7pt] (1.04,0) circle (0.145);
  \draw[blockedred, line width=0.7pt] (0.94,-0.10) -- (1.14,0.10);
  \draw[blockedred, line width=0.7pt] (0.94,0.10) -- (1.14,-0.10);
  \fill[black!75, rounded corners=1pt] (1.61,-0.13) rectangle (2.06,0.13);
\end{scope}}

\title{Similarity Gates Approve Reversals: A Validity Audit of Embedding-Cosine Thresholds in Agent Systems}
\author{Scott E. Frias\\ Eigenforma \textperiodcentered\ Freemind Labs\\ \texttt{scott@eigenforma.com}}
\date{August 4, 2026}

\begin{document}

\maketitle

\begin{abstract}
Agent frameworks ship quality gates that compare text blocks by embedding-cosine similarity and decide at a fixed cutoff. Deduplication filters, semantic caches, drift guards, and answer grader gates deploy to answer the question: ``Does this text still mean the same thing?'' But the score answers a \emph{different} question: ``How much did the \emph{wording} change?'' We audit this gate class as a measurement instrument. In the cases these gates exist to catch, the two can run in opposite ways. Many times, reversing an instruction is a single word edit, while agreement often rephrases a sentence. The consequence is a safety check that \emph{fires backwards}. The production drift guard we audited caught 0 of 56 meaning-breaking mutations, and one approved item, ``withhold the study drug'' $\rightarrow$ ``administer the study drug'', came in at cosine 0.9608. We observed five shipped operating points, and balanced accuracy across 90 configuration-threshold-task cells never exceeded 0.700 (median 0.525). The same confounder also corrupted evaluations. A naively built corpus inherits this confounder and can return an inverted verdict, with a decision AUROC exactly 0.000 in 13 of 18 configuration-task cells (at most 0.040 in all 18) against 0.440--0.815 for the same nine configurations under a balanced $2 \times 2$ design. Twice in the effort it captured our own headline claims. Obvious repairs fail: an encoder swap and an overlap-conditioned gate (0.750 in-sample, 0.533 held-out) land at chance on separately authored held-out data, and an NLI drop-in did no better. Embeddings do still bear hope here, as the strongest two of nine configurations separated reversal from paraphrase at matched overlap (AUROC 0.79--0.90), but only a matched-pair audit reveals the deployment regime. We release the corpus method, harness, and frozen results, and contend that scores gated this way measure the wrong thing. We believe a valid instrument is buildable.
\end{abstract}

\section{Introduction}

\begin{figure}[t]
\centering
\begin{tikzpicture}[line cap=round]
\begin{scope}
  \fill[black!80, rounded corners=2pt] (0,0.55) rectangle (1.5,1.35);
  \node[font=\scriptsize\ttfamily] at (0.75,0.30) {clinician};
  \draw[line width=0.8pt] (1.5,1.15) -- (5.5,1.15);
  \draw[line width=0.8pt] (1.5,0.75) -- (5.5,0.75);
  \draw[flowteal, line width=1.5pt] (1.6,0.95) -- (2.95,0.95);
  \draw[blockedred, line width=1.3pt] (3.35,0.95) circle (0.36);
  \draw[blockedred, line width=1.3pt] (3.10,0.70) -- (3.60,1.20);
  \draw[blockedred, line width=1.3pt] (3.10,1.20) -- (3.60,0.70);
  \node[font=\scriptsize\ttfamily, blockedred] at (3.35,1.62) {withhold};
  \draw[line width=1.0pt] (5.6,1.45) -- (5.6,0.45) -- (7.0,0.45) -- (7.0,1.45);
  \node[font=\scriptsize\ttfamily] at (6.3,0.20) {participant};
\end{scope}
\begin{scope}[shift={(8.3,0)}]
  \fill[black!80, rounded corners=2pt] (0,0.55) rectangle (1.5,1.35);
  \node[font=\scriptsize\ttfamily] at (0.75,0.30) {clinician};
  \draw[line width=0.8pt] (1.5,1.15) -- (5.5,1.15);
  \draw[line width=0.8pt] (1.5,0.75) -- (5.5,0.75);
  \draw[flowteal, line width=1.5pt] (1.6,0.95) -- (2.99,0.95);
  \draw[line width=1.2pt] (3.35,0.95) circle (0.36);
  \draw[line width=1.2pt] (3.13,1.17) -- (3.57,0.73);
  \node[font=\scriptsize\ttfamily] at (3.35,1.62) {administer};
  \draw[flowteal, line width=1.5pt, -{Stealth[length=5pt]}] (3.71,0.95) -- (5.45,0.95);
  \fill[flowteal, opacity=0.5] (5.66,0.48) rectangle (6.94,1.32);
  \draw[line width=1.0pt] (5.6,1.45) -- (5.6,0.45) -- (7.0,0.45) -- (7.0,1.45);
  \node[font=\scriptsize\ttfamily] at (6.3,0.20) {participant};
\end{scope}
\begin{scope}[shift={(1.1,-2.35)}, scale=0.82]
  \begin{scope}[opacity=1.0]
    \draw[line width=0.9pt, rounded corners=2pt] (0,0.55) rectangle (1.5,1.35);
    \draw[line width=0.9pt] (1.5,1.15) -- (5.5,1.15);
    \draw[line width=0.9pt] (1.5,0.75) -- (5.5,0.75);
    \draw[line width=0.9pt] (5.6,1.45) -- (5.6,0.45) -- (7.0,0.45) -- (7.0,1.45);
  \end{scope}
  \begin{scope}[shift={(0.12,-0.10)}, opacity=0.35]
    \draw[line width=0.9pt, rounded corners=2pt] (0,0.55) rectangle (1.5,1.35);
    \draw[line width=0.9pt] (1.5,1.15) -- (5.5,1.15);
    \draw[line width=0.9pt] (1.5,0.75) -- (5.5,0.75);
    \draw[line width=0.9pt] (5.6,1.45) -- (5.6,0.45) -- (7.0,0.45) -- (7.0,1.45);
  \end{scope}
  \node[font=\scriptsize\ttfamily, anchor=north] at (3.5,0.05) {silhouette overlap: what cosine reads};
\end{scope}
\begin{scope}[shift={(11.0,-1.85)}]
  \draw[line width=1.2pt] (0,0) circle (0.36);
  \draw[line width=1.2pt] (-0.22,0.22) -- (0.22,-0.22);
  \draw[blockedred, line width=1.2pt] (1.5,0) circle (0.36);
  \draw[blockedred, line width=1.2pt] (1.25,-0.25) -- (1.75,0.25);
  \draw[blockedred, line width=1.2pt] (1.25,0.25) -- (1.75,-0.25);
  \node[font=\scriptsize\ttfamily, anchor=north] at (0.75,-0.62) {the valves: where meaning lives};
\end{scope}
\node[draw=black!70, line width=0.6pt, inner sep=5pt, font=\small\ttfamily, anchor=north] at (7.9,-3.45)
  {cosine 0.9608 \quad guard fires below cosine 0.60 \quad it did not fire};
\end{tikzpicture}
\caption{The reversal, drawn. One flipped valve separates \emph{withhold} from \emph{administer}; the silhouettes nearly coincide, and the silhouette is all the gate reads. This held-out pair scored cosine 0.9608 against a firing line of cosine 0.60 on the production configuration (\S6). The guard did not fire.}
\label{fig:mechanism}
\end{figure}
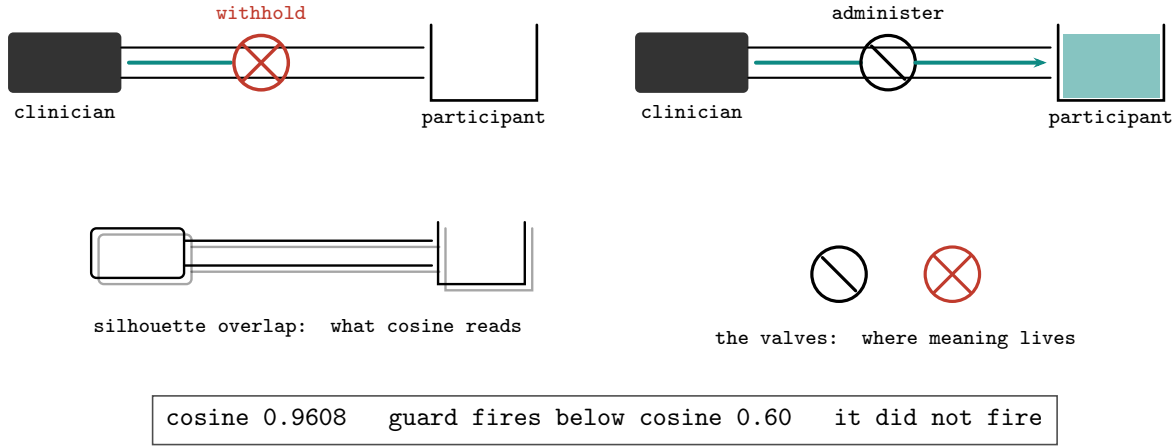

A family of shipped safety and quality checks assumes that embedding-cosine similarity measures agreement in meaning, where more accurately, these checks measure similarity in wording. In the pairs these checks exist to catch, wording and meaning are anti-correlated, and negating an instruction inserts a word or two and keeps the rest, whereas agreeing in fresh words replaces nearly all of them. A gate thresholding this score will pair a sentence to its own negation and treat the two as the same instruction. Further, the observable failures concentrated on one side, the \emph{pairing} side, where a reversal passes as the same instruction (\S6; Figure~\ref{fig:mechanism}). The four component test suites we audited contained no case that would reveal it.

In 2021, \hyperlink{bib:jacobs}{Jacobs and Wallach} located a broad class of system harms in the mismatch between a theoretical construct and its operationalization. \hyperlink{bib:li}{Li et al.\ (2026)} then name the specific mistake: the ``Proxy Presumption,'' treating geometric properties of embeddings as direct measures of a construct when the representation is an entangled mixture of that construct, containing confounders of topic, style, and authorship. Their confounder is, in our setting, lexical overlap; their discriminant-validity test is structurally what our factorial design performs. The component facts are also documented. Masked language models largely ignore negation (\hyperlink{bib:kassner}{Kassner \& Sch\"utze, 2020}; \hyperlink{bib:ettinger}{Ettinger, 2020}) and sentence-embedding similarity scores negated pairs among the \emph{highest}. \hyperlink{bib:blagec}{Blagec et al.\ (2019)} found negation and antonym subsets scoring above genuinely similar pairs on every model they tested, at average cosines of 0.93--0.999, and \hyperlink{bib:anschutz}{Ansch\"utz et al.\ (2023)} found the evaluation metrics learned from such models insensitive to negation. The evaluation layer's blindness to all of this is itself documented, in NLI benchmarks (\hyperlink{bib:hossain}{Hossain et al., 2020}) and sentence-encoder evaluation (\hyperlink{bib:chiang}{Chiang et al., 2023}).

What none of this literature audits is the deployed object itself: the \emph{control surface}, where an invalid instrument yields not a wrong estimate, but a safety check that fires backwards. Adjacent agent-systems work suggests that embedding-based behavioral measurement is \emph{less} sensitive to semantic shifts than human judgment (\hyperlink{bib:rath}{Rath, 2026}). \hyperlink{bib:zizzo}{Zizzo et al.\ (2025)} benchmark guardrails against a different threat model, adversarial prompts rather than unfaithful rewrites. \hyperlink{bib:wang}{Wang et al.\ (2026)} survey process-level accountability for agents (provenance, runtime guardrails, observability), and the validity of semantic-similarity instruments sits outside that survey's scope.

We contribute the audit and its artifacts:

\begin{enumerate}
\item \textbf{A corpus-design result.} The lexical--decision confounder in naturally authored evaluation corpora is structural (negation is an additive edit, paraphrase a substitutive one), so authorial intent cannot remove it, while matched generation can. A corpus reproducing the natural pattern made every encoder we tested look inverted, with a decision AUROC exactly 0.000 in 13 of 18 configuration-task cells, and at most 0.040 in all 18 (\S3). During this study, the same confounder captured two of our own claims (\S8).
\item \textbf{An audit instrument and what it shows.} A $2 \times 2$ factorial corpus (decision $\times$ lexical overlap, balanced per anchor, encoder-blind construction) separates what cosine measures from what the instrumented gates deploy it to measure. Further, the constructed lexical contrast was recovered as a positive control whose size depends on the constructed pole separation. We found stratified AUROC 0.975--1.000 across all nine configurations, and decision content on the same pairs spanned 0.440--0.815: heterogeneous across encoders, while at chance for the configuration deployed in the audited system (\S5).
\item \textbf{A deployed-system result.} The audited production drift guard caught 0 of 56 meaning-breaking mutations across two corpora, one authored under documented isolation. Its worst mutation classes were the operationally dangerous ones: weakening an obligation, or changing a quantity. Notably, not negation (\S6).
\item \textbf{Pre-registered negative results for the obvious repairs.} An encoder swap and an overlap-conditioned gate landed at chance on held-out data, and an NLI drop-in did no better (\S7).
\item \textbf{A released, offline, pinned harness and corpus method}, including a small prevalence survey of shipped operating points, and an audit that revealed a post-hoc hyperbolic projection pattern to be provably a no-op on normalized embeddings (\S9).
\end{enumerate}

The corpora are small (10 pairs per cell), single-annotator at present, and generated within one project's register; \S10 states these limits without minimizing them. The claims are demonstrated on our corpora and one deployed system, and indicated (not established) beyond them.

\section{The gate class under audit}

The pattern: embed two texts, compute cosine similarity, compare to a constant. Instances we audited directly or at their installed defaults:

\begin{center}
\small
\begin{tabular}{p{5.6cm}p{4.2cm}p{4.6cm}}
\toprule
component & shipped operating point & decision it makes \\
\midrule
production drift guard (audited system) & drift $= 1 - {}$cosine, threshold 0.4 & is this rewrite still the same instruction? \\
LlamaIndex \texttt{SemanticSimilarityEvaluator} & 0.8 & is this answer correct? \\
LangChain \texttt{EmbeddingsRedundantFilter} & 0.95 & is this document redundant? \\
semantic-cache hit logic (GPTCache-style \texttt{similarity\_threshold}) & 0.8 & may this cached answer be reused? \\
\bottomrule
\end{tabular}
\end{center}

Vendors and practitioners publish explicit operating points for the pattern (e.g.\ a cosine-distance cut of 0.15 recommended as indicating meaningful drift, and rolling baselines firing at fixed deviations). The exact sources are recorded in the released prevalence survey. A keyword survey returned 36 candidate gate sites, and of the 11 we opened and read, 9 were real cosine-threshold gates with observed thresholds spanning 0.30--0.95, including one that auto-merges entities at 0.95. We report opened-and-read counts only, as a prevalence \emph{rate} would require a sampling frame and a two-coder protocol that we have not run.

\section{The confounder, and what a naive evaluation returns}

\textbf{Mechanism.} Negation and its relatives (scope inversion, modal weakening, quantity change) are additive or minimal edits, as they insert or swap a token and keep the rest. Restating a decision faithfully is substitutive and \emph{replaces} tokens. The asymmetry is linguistic before it is statistical (negation is the marked member of the affirmative--negative pair, expressed by operating on an affirmative base rather than by composing a fresh sentence; \hyperlink{bib:horn}{Horn, 1989}), so it is a property of how people write, not of any one corpus-builder's habits. In our corpora, the natural reversal sits at token-Jaccard $\approx 0.72$ with its anchor, while the faithful full rewrite sits at $\approx 0.06$. A corpus authored the natural way therefore couples lexical overlap to the class label. In our corpora the coupling ran strongly enough to invert the verdict (which actually helped spur this discovery), though its strength varied by corpus and authoring context (\S7). Our first hand-authored corpus failed its own manipulation check on exactly this (within-stratum overlap 0.721 vs 0.520, permutation $p = 0.014$). We generated the released corpus with per-anchor, encoder-blind overlap matching instead, and it passed (per-stratum balance $p$ = 0.108--0.980). PAWS (\hyperlink{bib:zhang}{Zhang et al., 2019}) established the resource-side half of this problem (models fail on high-overlap non-paraphrases until trained on them), but an adversarial paraphrase set supplies only one cell of the design a gate audit needs; it seems the factorial requires all four, including the low-overlap \emph{agreements} that no such set contains.

\textbf{What the naive design returns.} Simulating the natural authoring pattern inside our balanced corpus (lexically close opposites versus lexically distant agreements, 10 vs 10 per task) yielded decision AUROC of exactly 0.000 in 13 of 18 configuration-task cells and $\le 0.040$ in all 18. Under an unrestricted permutation null the perfect separations carry $p = 1.08 \times 10^{-5}$ each. The anchor-matched design violates that null's exchangeability, and the honest paired-design floor is $p \approx 0.002$ ($1.95 \times 10^{-3}$), still decisive. Under the balanced design the same encoder configurations spanned stratified decision AUROC 0.440--0.815, measured over all 40 pairs per task where the naive comparison sees 20. The stronger half of the field reached 0.73--0.815, and the production configuration stayed at chance. The inversion is a property of the corpus design, realized through the encoders' lexical dominance. Remove either and it disappears.

\textbf{It is not a toy.} The pattern reproduced on both of our pre-existing corpora: a hypothesis-distinctness set (cosine AUROC 0.000, $n = 8$ vs 8, exact $p = 1.6 \times 10^{-4}$) and the drift guard's held-out mutation corpus (0.267, $n = 30$ vs 10, 95\% CI [0.110, 0.438], excluding chance in the \emph{wrong} direction). Their sibling splits (0.25 at $n = 4$ vs 4; 0.285 in-sample) pointed the same way with CIs touching chance. We cite all four so the selection is visible. This measurement artifact is a plausible origin of the folklore that embeddings simply cannot represent negation: measured naively, they look worse than blind. They look inverted. The published record is consistent with this reading. The corpus-level inversions measured in prior work, negation and antonym subsets scoring above highly similar pairs at near-ceiling average cosines (\hyperlink{bib:blagec}{Blagec et al., 2019}), were observed on naturally assembled subsets, which is exactly where the confounder mechanism predicts them.

\textbf{Ecological structure vs.\ representation.} In natural authoring inputs, minimal-edit reversals share high lexical overlap with their anchor (mean Jaccard $\approx 0.715$); faithful full restatements share low overlap ($\approx 0.062$). Compared unmatched (9 configurations $\times$ 20 anchors), reversal appeared to move the embedding less than restatement in 180 of 180 cases (2.6--10.3$\times$). The adversarial review of \S8 established that this ratio is a property of the anti-matching, not an intrinsic blindness of the representations. With lexical overlap held constant, the direction nullified or inverted. Moreover, reversal moved the embedding \emph{more} than restatement in 29 of 36 configuration-task-stratum comparisons in the frozen factorial results. We report the unmatched comparison because deployment inputs are shaped exactly like it.

\section{The instrument}

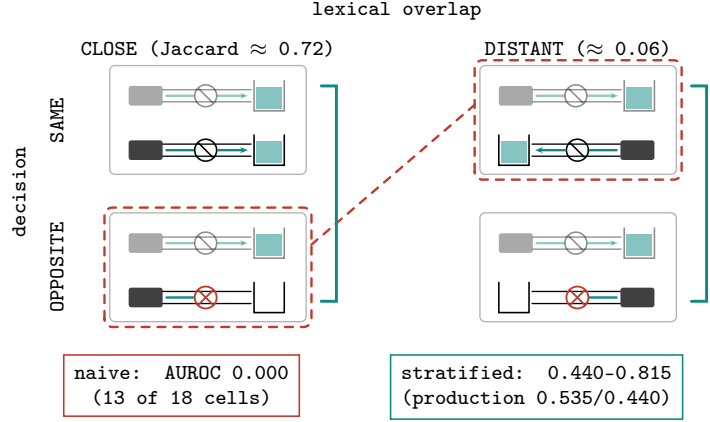
\begin{figure}[t]
\centering
\begin{tikzpicture}[line cap=round]
\node[font=\scriptsize\ttfamily] at (3.55,1.55) {lexical overlap};
\node[font=\scriptsize\ttfamily] at (1.05,1.05) {CLOSE (Jaccard $\approx$ 0.72)};
\node[font=\scriptsize\ttfamily] at (5.95,1.05) {DISTANT ($\approx$ 0.06)};
\node[font=\scriptsize\ttfamily, rotate=90] at (-1.45,-0.85) {decision};
\node[font=\scriptsize\ttfamily, rotate=90] at (-0.95,0.1) {SAME};
\node[font=\scriptsize\ttfamily, rotate=90] at (-0.95,-1.85) {OPPOSITE};
\draw[black!30, line width=0.5pt, rounded corners=2pt] (-0.25,-0.62) rectangle (2.35,0.82);
\draw[black!30, line width=0.5pt, rounded corners=2pt] (4.65,-0.62) rectangle (7.25,0.82);
\draw[black!30, line width=0.5pt, rounded corners=2pt] (-0.25,-2.57) rectangle (2.35,-1.13);
\draw[black!30, line width=0.5pt, rounded corners=2pt] (4.65,-2.57) rectangle (7.25,-1.13);
\FSopen{0}{0.42}{0.45}
\FSopen{0}{-0.30}{1}
\FSopen{0}{-1.53}{0.45}
\FScross{0}{-2.25}{1}
\FSopen{4.9}{0.42}{0.45}
\FSopenM{4.9}{-0.30}{1}
\FSopen{4.9}{-1.53}{0.45}
\FScrossM{4.9}{-2.25}{1}
\draw[blockedred, dashed, line width=0.9pt, rounded corners=3pt] (-0.32,-2.64) rectangle (2.42,-1.06);
\draw[blockedred, dashed, line width=0.9pt, rounded corners=3pt] (4.58,-0.69) rectangle (7.32,0.89);
\draw[blockedred, dashed, line width=0.9pt] (2.42,-1.55) -- (4.58,0.3);
\draw[flowteal, line width=1.1pt] (2.55,0.55) -- (2.75,0.55) -- (2.75,-2.3) -- (2.55,-2.3);
\draw[flowteal, line width=1.1pt] (7.45,0.55) -- (7.65,0.55) -- (7.65,-2.3) -- (7.45,-2.3);
\node[draw=blockedred, line width=0.6pt, inner sep=4pt, font=\scriptsize\ttfamily, anchor=north, align=center] at (0.7,-3.0)
  {naive: AUROC 0.000\\(13 of 18 cells)};
\node[draw=flowteal, line width=0.6pt, inner sep=4pt, font=\scriptsize\ttfamily, anchor=north, align=center] at (5.4,-3.0)
  {stratified: 0.440--0.815\\(production 0.535/0.440)};
\end{tikzpicture}
\caption{The instrument. Every cell pairs the same anchor (gray) with a variant (ink); columns fix lexical overlap, rows fix the decision. The naive comparison (dashed) contrasts close opposites with distant agreements and reads the confounder: decision AUROC exactly 0.000 in 13 of 18 configuration-task cells. The stratified comparison (solid) holds overlap fixed and reads the decision: AUROC 0.440--0.815 across nine configurations, with the production configuration at 0.535/0.440 (\S3, \S5).}
\label{fig:instrument}
\end{figure}

A $2 \times 2$ factorial corpus per task, decision (same/opposite; faithful/violation) $\times$ lexical overlap (close/distant), with all four cells sharing each anchor (Figure~\ref{fig:instrument}). In all, 10 anchors per task, 10 pairs per cell, two tasks (hypothesis distinctness; constraint survival), and 80 pairs total. Overlap balanced within decision classes per anchor, by construction, with no encoder loaded. Nine encoder configurations from seven checkpoints (a prefix-and-dimension pair for nomic; two prefix variants for e5), including the audited system's exact production path (\texttt{nomic-embed-text-v1.5} at MRL-256, routed through the shipped embedding router). All runs are offline against pinned local models; results are frozen JSON. The corpus, harness, and per-pair provenance ship as the artifact.

We used two measurements per configuration and task: the \textbf{decision axis} (same vs opposite decision, within lexical strata, stratified AUROC) and the \textbf{lexical axis} (close vs distant overlap, within decision classes) acting as a positive control. The design is a discriminant-validity test, in the sense the term has carried since \hyperlink{bib:campbell}{Campbell and Fiske (1959)}: the measure must track the construct the gate actually deploys it for and fail to track the \emph{confounder} it is suspected of measuring instead, with both contrasts instrumented on the same matched pairs.

\section{What the valid instrument shows}

\textbf{The positive control.} All nine configurations recovered the constructed lexical contrast at stratified AUROC 0.975--1.000 on both tasks. The instrument and encoders work, so what follows is not noise. That figure depends on how far apart we placed the overlap poles ($\approx 0.64/0.68$ Jaccard separation by task). Narrowing to a within-stratum median split ($\approx$ 0.08--0.16 separation) yielded 0.62--0.93 (medians $\approx 0.72/0.74$), and cosine tracked Jaccard continuously within strata (median Spearman 0.35/0.43). Cosine is a graded surface-form meter, not a threshold detector of it.

\textbf{The decision axis is heterogeneous, and the deployed configuration is at chance.} Stratified decision AUROC spanned 0.490--0.808 (distinctness, median 0.740) and 0.440--0.815 (constraint, median 0.728). The production configuration scored 0.535/0.440, with its constraint CLOSE stratum point-estimate inverted at 0.32 (CI [0.09, 0.57], not excluding chance). Restoring full dimensionality under the documented symmetric prefix scored 0.490/0.550, so MRL truncation alone is not the cause. The strongest configurations \emph{can} make the distinction the gate needs. On the direct contrast that a gate exists to make (total reversal versus paraphrase at matched $\approx 0.72$ overlap), \texttt{bge} reached AUROC 0.900 (CI [0.73, 1.00]), \texttt{mxbai} 0.895, the \texttt{e5} variants 0.85--0.87, and \texttt{gte} 0.815 on the distinctness task. On the constraint task, the same leading pair scored 0.790 (\texttt{bge}) and 0.820 (\texttt{mxbai}). The strongest two configurations therefore spanned AUROC 0.79--0.90 across both tasks, and 8 of the 18 matched-overlap contrasts excluded chance upward. Encoder choice moved the decision axis (spreads of 0.318/0.375 in stratified AUROC by task) far more than any threshold policy below recovered.

\textbf{No single threshold serves even the strong encoders.} The optimal cut moved 0.11--0.56 cosine between lexical strata (median 0.264, $n = 18$ configuration-tasks), and the gate never observes the stratum. Forcing one global cut cost 0.05--0.15 balanced accuracy against the per-stratum oracle. At the five shipped operating points (0.30, 0.40, 0.80, 0.85, 0.95, all on the cosine-similarity scale), balanced accuracy across all 90 configuration-threshold-task cells was 0.450--0.700 (median 0.525). (Provenance: 0.30 and 0.40 read from surveyed memory-gate code, 0.80 from framework defaults, 0.85 from the vendor drift cut, 0.95 from the redundancy filter. The audited guard's threshold of 0.4 lives on the drift scale, i.e.\ cosine 0.60, and is evaluated separately in \S6.) The best any configuration achieved at any shipped threshold was 0.525--0.700, and the single best cell anywhere was 0.700.

\textbf{Mechanism does not rescue policy.} A continuous score with no threshold (mean class gap 0.0009--0.087), top-k paraphrase mining (wrong-pair rate 0.2--0.4), and an adaptive-threshold chunker (0 of 20 meaning flips detected, all nine encoders) failed the same way, because each inherits the same score.

\section{The deployed system}

The audited drift guard (production encoder, drift cut 0.4, i.e.\ cosine 0.60) fired on 0 of 56 meaning-breaking mutations. The mutations span the in-sample corpus (26 mutations, 5 faithful controls) and a held-out corpus (30, 10) authored under documented isolation. A separate authoring session saw neither our corpora nor our code. (\emph{Note}: the isolation is an isolated context, not an independent human; see \S10.) Every mutation scored cosine 0.832--0.9997, meaning the threshold is unreachable \emph{by construction}, and no operating point in the deployed direction exceeded balanced accuracy 0.58, even when tuned on the test data itself (0.573 in-sample, 0.550 held-out). The guard also fired on none of the 15 faithful controls (cosines 0.869--0.993): it approved everything it saw. By mutation class, the guard scored worst on operationally dangerous classes: modal downgrade (\texttt{must} $\rightarrow$ \texttt{should}; AUROC 0.000 in-sample, 0.125 held-out) and quantity/unit drift (0.160/0.067). Modal downgrade is hedging by another name, and hedging patterns are an explicit target of recent negation-robust training data (\hyperlink{bib:truong}{Truong et al., 2025}). The classes nearest chance were constraint deletion (0.480/0.467) and negation (0.433/0.442), ironically the class folklore blames. One held-out specimen: \emph{``Withhold the study drug from any participant who reports chest tightness''} $\rightarrow$ \emph{``Administer the study drug\ldots''} scored cosine 0.9608. The guard \textbf{did not fire}. Clinical NLP has treated negation as demanding dedicated machinery for two decades. NegEx (\hyperlink{bib:chapman}{Chapman et al., 2001}) exists precisely because surface similarity does not carry polarity. This gate class ships with no equivalent.

At installed defaults on our corpora the framework components behaved consistently with this. The answer-grader accepted wrong answers at rates of 0.425--0.85 (Youden's $J$ 0.025--0.175 across nine configurations), and the redundancy filter silently dropped the disagreeing document 20--90\% of the time. All four framework test suites we audited contained no polarity case (no test in which two texts share surface form and differ in decision), which we take to be \emph{why} this gate class ships: the failure mode is absent from the tests. This is the component-level echo of a benchmark-level finding, reproduced at the point where the score has become a control decision, and \hyperlink{bib:hossain}{Hossain et al.\ (2020)} showed that NLI corpora underrepresent negation so thoroughly that a model can ignore it, and still score well.

\section{The repairs fail out of domain}

We pre-registered two repairs a practitioner would try first (bars, predictions, and fit/evaluation separation committed before the run), fitting every parameter on the factorial corpus and evaluating frozen on the two drift-guard corpora (different texts, separately authored). Those two corpora are the table's evaluation columns. The third row reports a frozen prior result on the same held-out corpus.

\begin{center}
\small
\begin{tabular}{p{4.6cm}cc p{5.2cm}}
\toprule
repair & in-sample & held-out & outcome \\
\midrule
swap to the best-auditing encoder (\texttt{mxbai}), keep the gate & 0.485 & 0.433 & failed, as predicted; in the natural input regime the confounder defeats any single-cosine policy regardless of encoder \\
condition the gate on the confounder: logistic on (cosine, token-Jaccard) & \textbf{0.750} & \textbf{0.533} & failed its held-out bar ($\ge 0.70$); this was the primary arm and our pre-registered prediction said it would pass \\
NLI cross-encoder drop-in (frozen prior result; AUROC) & 0.831 & 0.533 (CI [0.36, 0.71]) & not a repair on this evidence \\
\bottomrule
\end{tabular}
\end{center}

The conditioned gate's failure restates the paper's thesis. In-sample faithful controls sat at Jaccard 0.11--0.50 (mean 0.33), below the mutation mean of $\approx 0.65$ (though the ranges do overlap at the tails) and the held-out corpus's faithful controls spanned 0.06--0.79. The wording--meaning correlation is not a fixable covariate but an unstable one, as it shifted between two corpora, produced in different authoring sessions, under different protocols. A team that evaluated the conditioned gate on its own corpus would have shipped it at 0.750 and fielded a coin-flip. Evaluation across an authoring change is what catches this (\S11).

We do not conclude the gate class is unrepairable. We conclude only that nothing calibration-shaped that we tested survived an author change, and that the surviving directions are architectural. The first is pair-level judging, scoped to the mutation classes a judge demonstrably catches. The second is narrowing the gate's scope of authority to the surface-form property cosine actually measures. Both are directions, not demonstrations. A literature of negation-aware representations exists to build on, such as encoders trained on LLM-distilled negation and hedging data (\hyperlink{bib:truong}{Truong et al., 2025}), negation-focused similarity architectures (\hyperlink{bib:rohit}{Rohit et al., 2026}), and negation-aware metrics (\hyperlink{bib:anschutz}{Ansch\"utz et al., 2023}). Our held-out result argues that any candidate should be evaluated across an author change before it is trusted in a gate.

\section{The confounder captured its own investigators}

Twice during this study, the confounder produced a result we briefly held as a headline claim. A reversal-versus-restatement comparison, matched on topic by sharing an anchor, showed reversals moving embeddings 2.6--10.3$\times$ less than restatements in 180 of 180 cases. It was also anti-matched on lexical overlap by construction (\S3), comparing high-overlap reversals (Jaccard $\approx 0.715$) against low-overlap restatements ($\approx 0.062$): the very defect the corpus was built to expose. A derived magnitude-equivalence claim fell the same way.

A deep adversarial review of our own claims removed both, and its scoring rules were committed before any result was seen. Most of our pre-registered predictions about which claims were safest were wrong, and the repair study's primary prediction was wrong as well (\S7). We report this not as method narrative but as evidence about the phenomenon. In our case, the confounder captured careful investigators actively watching for it. Vigilance did not protect us, but the de-confounded design did. The claims retained here are the less profound, but they are the ones that survived attack. The review did not cover everything (\S10 lists what went unassaulted).

\section{A note on post-hoc hyperbolic projection}

Hyperbolic representation learning is legitimate and effective when the representation is trained in the hyperbolic space (\hyperlink{bib:nickel}{Nickel \& Kiela, 2017}). One audited pipeline instead projected L2-normalized embeddings into a Poincar\'e ball post-hoc and ranked by geodesic distance. On unit-norm inputs this is provably a no-op for any curvature parameter $c > 0$, as all points land at one radius, and geodesic distance is strictly monotone in the cosine the pipeline already possessed.

Formally, for the Poincar\'e exponential map $\exp_0^c(v) = \frac{\tanh(\sqrt{c}\,\|v\|)}{\sqrt{c}\,\|v\|}\, v$ (the curvature-$c$ formalism of \hyperlink{bib:ganea}{Ganea et al., 2018}), every unit vector ($\|u\| = \|v\| = 1$) lands at radius $r(c) = \frac{\tanh(\sqrt{c})}{\sqrt{c}}$, making $1 - c\, r(c)^2 = \operatorname{sech}^2(\sqrt{c})$, the factor that carries the entire hyperbolic content of the metric, a constant. The geodesic distance collapses to
\[
d_c(\exp_0^c(u), \exp_0^c(v)) = \frac{1}{\sqrt{c}} \operatorname{arcosh}\!\left(1 + \tfrac{1}{2}\sinh^2(2\sqrt{c})\, \|u - v\|^2\right),
\]
a strictly increasing function of the Euclidean distance $\|u - v\| = \sqrt{2(1 - \cos(u,v))}$, hence strictly decreasing in cosine.

The algebra is known. \hyperlink{bib:moreira}{Moreira et al.\ (2024)} show that constant-radius hyperbolic embeddings behave as Euclidean ones on a sphere. We contribute the audit herein as a closed form checked to $1.3 \times 10^{-15}$ across 400 unit-norm vectors at the encoder's dimensionality (a conservative stand-in, since the proposition depends only on norms), rankings identical vector-for-vector, against a from-description re-implementation of the third-party functions (the original code was read but not redistributed, for license reasons). A very small survey of $\sim 60$ repositories using exponential maps found most doing legitimate hyperbolic geometry on trainable representations (though we did confirm one instance of the no-op pattern in the wild).

\section{Limitations}

(i) The factorial corpus is small (10 pairs per cell), so most per-stratum decision CIs individually touch chance. The field-level statements in \S5 are what the data support. (ii) Labels are single-annotator, though independent-annotator slots exist in the released corpus and $\kappa$ is unfilled at this writing. (iii) One deployed system was audited end-to-end, and the framework probes are at installed defaults on our corpora, not in-situ deployments. (iv) The corpora are generated within one project's register (short imperative software/clinical instructions) by one model under one operator. The corpus-design finding of \S3 is our reason to prefer generated matched pairs over hand-authoring, and it is also a real external-validity limit. (v) Nine configurations span seven checkpoints, and the paired variants are sensitivity controls, not independent evidence. (vi) The prevalence observations are opened-and-read counts, not rates. (vii) LLM agents under contamination controls (not independent humans) performed the adversarial review of our claims (\S8), and it targeted the headline claims only: the no-threshold, test-suite, mechanism-variance, and projection results were \emph{not} attack targets, and carry no adversarial clean bill. Human replication has not yet occurred.

\section{What a practitioner should do, on this evidence}

\begin{enumerate}
\item \textbf{Treat a cosine threshold over text pairs as a \emph{surface-form} gate.} If the property you enforce is decision agreement, do not rely on it alone at any threshold.
\item \textbf{Before shipping such a gate, audit your configuration with matched pairs.} (The released harness does this offline in one command.) Expect encoder regime to dominate. Be aware that some configurations can see reversals at matched overlap, others cannot, and the deployed one in our audit could not.
\item \textbf{Add polarity cases to component test suites:} pairs that share wording and flip decision (negation, scope, modal strength, quantity), and pairs that share decision and share no wording. Four of four suites we audited had none.
\item \textbf{Do not evaluate the gate, or its repair, only on corpora written by the people who built it.} Our conditioned repair looked fixed in-domain (0.750) and sat at chance on the separately authored held-out corpus (0.533). The failure mode of the evaluation is the same as the failure mode of the gate.
\end{enumerate}

\section{Conclusion}

The score at the center of a widely shipped gate class tracks \emph{wording} far more strongly than \emph{meaning}. The authoring patterns we measured made the two anti-correlated in precisely the pairs that those gates exist to catch. The shipped tests we audited cannot see it, the natural evaluation design reports its opposite, and the cheap repairs did not survive a change of authoring context. None of this argues that embeddings are useless, as the strongest encoders separated reversal from paraphrase when the comparison was made valid. The evidence argues that validity must be \emph{measured}, per configuration, with a de-confounded instrument, before a similarity score is deployed as a safety decision. We release the instrument. It is cheap, it runs offline, and, as we found, it catches exactly the errors that appear as strong results.

\section*{Artifact availability}

The corpus, corpus-construction method, audit harnesses, frozen results, and the prevalence survey's source record ship as a single offline artifact; every experiment reproduces with one command against pinned local models. The closed form of \S9 is machine-checked in the artifact against the general Poincar\'e distance across multiple curvatures, and the 29-of-36 matched-comparison count of \S3 recomputes from the frozen per-cell means. The artifact is public at \url{https://github.com/eigenforma/polaritycheck}, released as v1.0.0 and archived at DOI \href{https://doi.org/10.5281/zenodo.21796531}{10.5281/zenodo.21796531}.

\section*{References}

\bibentry{\hypertarget{bib:anschutz}{}Ansch\"utz, M., Lozano, D. M., \& Groh, G. (2023). \emph{This is not correct! Negation-aware evaluation of language generation systems.} INLG 2023. arXiv:2307.13989.}
\bibentry{\hypertarget{bib:blagec}{}Blagec, K., Xu, H., Agibetov, A., \& Samwald, M. (2019). \emph{Neural sentence embedding models for semantic similarity estimation in the biomedical domain.} BMC Bioinformatics, 20, 178. arXiv:2110.15708.}
\bibentry{\hypertarget{bib:campbell}{}Campbell, D. T., \& Fiske, D. W. (1959). \emph{Convergent and discriminant validation by the multitrait-multimethod matrix.} Psychological Bulletin, 56(2), 81--105.}
\bibentry{\hypertarget{bib:chapman}{}Chapman, W. W., Bridewell, W., Hanbury, P., Cooper, G. F., \& Buchanan, B. G. (2001). \emph{A simple algorithm for identifying negated findings and diseases in discharge summaries.} Journal of Biomedical Informatics, 34(5), 301--310.}
\bibentry{\hypertarget{bib:chiang}{}Chiang, C.-H., Chuang, Y.-S., Glass, J., \& Lee, H.-y. (2023). \emph{Revealing the blind spot of sentence encoder evaluation by HEROS.} Repl4NLP Workshop, ACL 2023. arXiv:2306.05083.}
\bibentry{\hypertarget{bib:ettinger}{}Ettinger, A. (2020). \emph{What BERT is not: Lessons from a new suite of psycholinguistic diagnostics for language models.} Transactions of the ACL, 8. arXiv:1907.13528.}
\bibentry{\hypertarget{bib:ganea}{}Ganea, O.-E., B\'ecigneul, G., \& Hofmann, T. (2018). \emph{Hyperbolic neural networks.} NeurIPS 2018. arXiv:1805.09112.}
\bibentry{\hypertarget{bib:horn}{}Horn, L. R. (1989). \emph{A Natural History of Negation.} University of Chicago Press.}
\bibentry{\hypertarget{bib:hossain}{}Hossain, M. M., Kovatchev, V., Dutta, P., Kao, T., Wei, E., \& Blanco, E. (2020). \emph{An analysis of natural language inference benchmarks through the lens of negation.} EMNLP 2020, 9106--9118.}
\bibentry{\hypertarget{bib:jacobs}{}Jacobs, A. Z., \& Wallach, H. (2021). \emph{Measurement and fairness.} FAccT 2021. arXiv:1912.05511.}
\bibentry{\hypertarget{bib:kassner}{}Kassner, N., \& Sch\"utze, H. (2020). \emph{Negated and misprimed probes for pretrained language models: Birds can talk, but cannot fly.} ACL 2020. arXiv:1911.03343.}
\bibentry{\hypertarget{bib:li}{}Li, B., Yu, T., Koa, K. J. L., \& Huang, K.-W. (2026). \emph{The Proxy Presumption: From semantic embeddings to valid social measures.} ACL 2026. arXiv:2605.07409.}
\bibentry{\hypertarget{bib:moreira}{}Moreira, G., Marques, M., Costeira, J. P., \& Hauptmann, A. (2024). \emph{Hyperbolic vs Euclidean embeddings in few-shot learning: Two sides of the same coin.} WACV 2024. arXiv:2309.10013.}
\bibentry{\hypertarget{bib:nickel}{}Nickel, M., \& Kiela, D. (2017). \emph{Poincar\'e embeddings for learning hierarchical representations.} NeurIPS 2017. arXiv:1705.08039.}
\bibentry{\hypertarget{bib:rath}{}Rath, A. (2026). \emph{Agent drift: Quantifying behavioral degradation in multi-agent LLM systems over extended interactions.} arXiv:2601.04170.}
\bibentry{\hypertarget{bib:rohit}{}Rohit, M., Jeganathan, L., Ummity, S. R., Janaki Meena, M., \& Balabaskaran, J. (2026). \emph{Computation of sentence similarity score through hybrid deep learning with a special focus on negation sentence.} Scientific Reports. doi:10.1038/s41598-025-34084-2.}
\bibentry{\hypertarget{bib:truong}{}Truong, T. H., Verspoor, K., Cohn, T., \& Baldwin, T. (2025). \emph{Learning robust negation text representations.} arXiv:2507.12782.}
\bibentry{\hypertarget{bib:wang}{}Wang, Y., Zhang, J., Cai, T., Liu, Z., Sun, Q., Sun, Z., Wu, Z., Dong, M., Zheng, M., Yin, X., \& Zhu, Y. (2026). \emph{From agent traces to trust: A survey of evidence tracing and execution provenance in LLM agents.} arXiv:2606.04990.}
\bibentry{\hypertarget{bib:zhang}{}Zhang, Y., Baldridge, J., \& He, L. (2019). \emph{PAWS: Paraphrase adversaries from word scrambling.} NAACL 2019. arXiv:1904.01130.}
\bibentry{\hypertarget{bib:zizzo}{}Zizzo, G., Cornacchia, G., Fraser, K., Hameed, M. Z., Rawat, A., Buesser, B., Purcell, M., Chen, P.-Y., Sattigeri, P., \& Varshney, K. (2025). \emph{Adversarial prompt evaluation: Systematic benchmarking of guardrails against prompt input attacks on LLMs.} Safe Generative AI Workshop, NeurIPS 2024. arXiv:2502.15427.}

\paragraph{Audited software artifacts.} \texttt{llama-index-core} (\texttt{SemanticSimilarityEvaluator}), \texttt{langchain} (\texttt{EmbeddingsRedundantFilter}), and the nine encoder configurations of \S4, at the installed versions pinned in the released harness; thresholds were read from the installed packages, not from documentation.

\end{document}